\documentclass{article}

\usepackage{amsmath} 
\usepackage{array}

    \PassOptionsToPackage{numbers, compress}{natbib}

\usepackage[dblblindworkshop, final]{neurips_2025}

\usepackage[utf8]{inputenc} 
\usepackage[T1]{fontenc}    
\usepackage{hyperref}       
\usepackage{url}            
\usepackage{booktabs}       
\usepackage{amsfonts}       
\usepackage{nicefrac}       
\usepackage{microtype}      
\usepackage{textgreek}      
\usepackage{xcolor}         
\usepackage{tikz}          
\usetikzlibrary{shapes.geometric, arrows, positioning, fit, backgrounds}

\usepackage{listings}
\lstdefinestyle{markdown}{
  breaklines=true,
  breakatwhitespace=true,
  columns=fullflexible,
  keepspaces=true,
  frame=single,
  xleftmargin=2em, xrightmargin=2em,
  xleftmargin=1em, xrightmargin=1em,
  aboveskip=1\baselineskip, belowskip=1\baselineskip
}

\tikzstyle{block} = [rectangle, draw, fill=blue!20, 
    text width=5em, text centered, rounded corners, minimum height=4em]
\tikzstyle{line} = [draw, -latex']

\title{AgentChangeBench: A Multi-Dimensional Evaluation Framework for Goal-Shift Robustness in Conversational AI}

\author{%
  Manik Rana \\
  Puch AI \\
  manik@puch.ai \\
  \And
  Calissa Man \\
  Algoverse \\
  calissaman@gmail.com \\
  \And
  Anotida Expected \\
  Algoverse \\
  anoexpected@gmail.com \\
  \And
  Jeffrey Paine \\
  Algoverse \\
  jeffrey.paine@gmail.com \\
  \And
  Kevin Zhu \\
  Algoverse \\
  kevin@algoverseacademy.com \\
  \And
  Vasu Sharma \\
  Algoverse \\
  sharma.vasu55@gmail.com \\
  \And
  Sunishchal Dev \\
  Algoverse \\
  dev@algoverseairesearch.org \\
  \And
  Ahan M R$^{\dagger}$ \\
  Microsoft \\
  ahanmr@microsoft.com \\
}

\begin{document}

\maketitle
\renewcommand{\thefootnote}{$\dagger$}
\footnotetext{Project lead}
\begin{abstract}
Goal changes are a defining feature of real world multi-turn interactions, yet current agent benchmarks primarily evaluate static objectives or one-shot tool use. We introduce \textbf{AgentChangeBench}, a benchmark explicitly designed to measure how tool augmented language model agents adapt to mid dialogue goal shifts across three enterprise domains. Our framework formalizes evaluation through four complementary metrics: Task Success Rate (TSR) for effectiveness, Tool Use Efficiency (TUE) for reliability, Tool Call Redundancy Rate (TCRR) for wasted effort, and Goal-Shift Recovery Time (GSRT) for adaptation latency. AgentChangeBench comprises 2,835 task sequences and five user personas, each designed to trigger realistic shift points in ongoing workflows. Using this setup, we evaluate several frontier models and uncover sharp contrasts obscured by traditional pass@k scores: for example, GPT-4o reaches 92.2\% recovery on airline booking shifts while Gemini collapses to 48.6\%, and retail tasks show near perfect parameter validity yet redundancy rates above 80\%, revealing major inefficiencies. These findings demonstrate that high raw accuracy does not imply robustness under dynamic goals, and that explicit measurement of recovery time and redundancy is essential. AgentChangeBench establishes a reproducible testbed for diagnosing and improving agent resilience in realistic enterprise settings.

\end{abstract}


\begin{figure}[h!]
  \centering
  \includegraphics[width=0.5\linewidth]{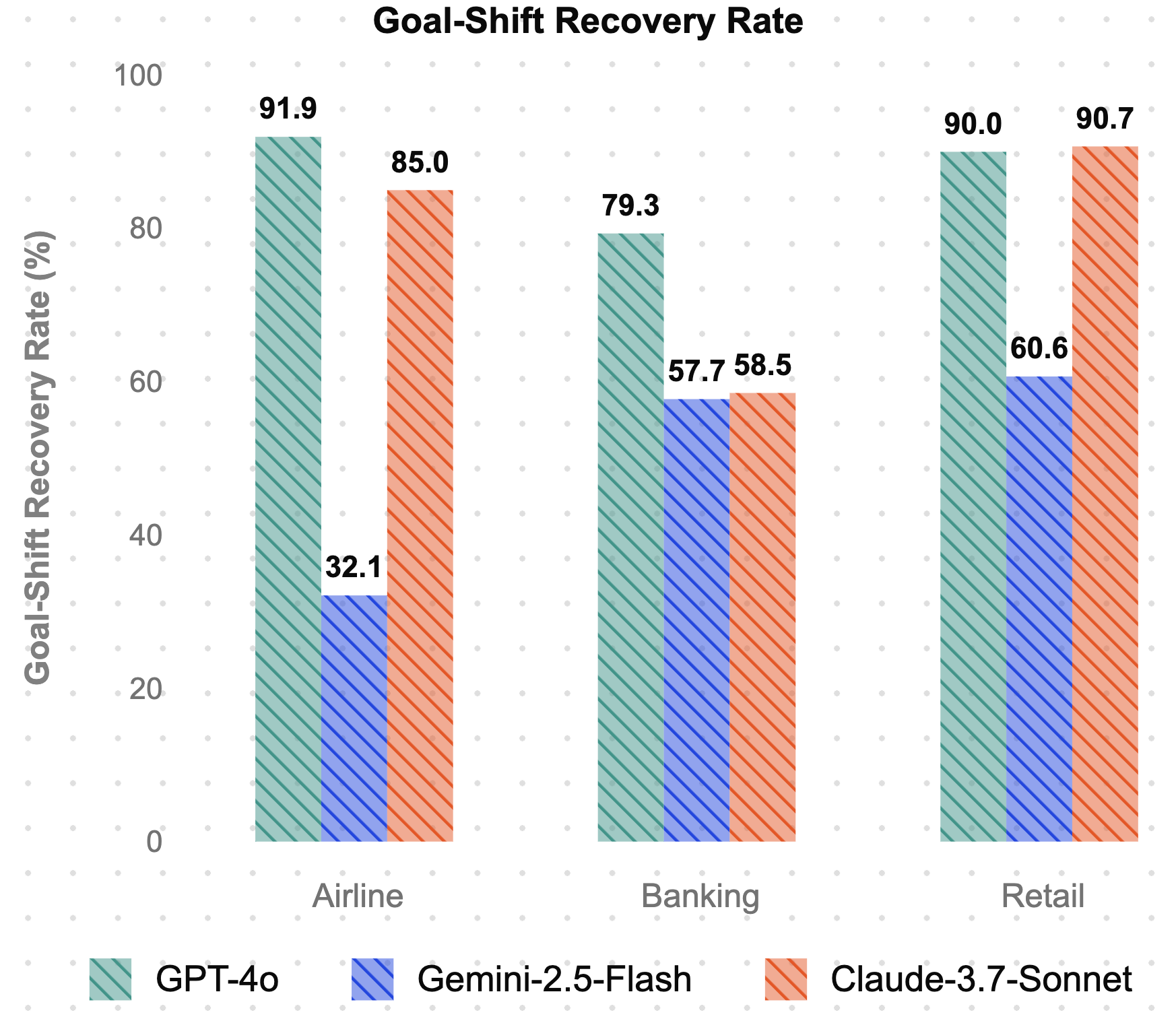}
\caption{\textbf{GSRT by model and domain.} Turns after a user goal shift to acknowledgment, first relevant tool call, and completion, lower is better. Aggregated over all tasks and personas in banking, retail, and airline for GPT-4o, Claude-3.7-Sonnet, and Gemini-2.5-Flash; labels show recovery rate, percent of shifts acknowledged without human transfer. Source: Appendix Tables~\ref{tab:efficiency_full},~\ref{tab:gs_new_only}.}

\end{figure}

\section{Introduction}

Large Language Models (LLMs) have rapidly advanced as conversational agents capable of reasoning, tool use, and multi-turn interaction across diverse domains. However, most existing benchmarks for evaluating LLM-as-agent performance assume that user goals remain fixed throughout a conversation. This assumption oversimplifies real-world deployments, where users frequently re-prioritize tasks, introduce new constraints, or shift objectives mid-dialogue. For example, a banking customer may begin by authenticating their identity, then pivot to reviewing transactions, and finally escalate to disputing a fraudulent charge, all within the same interaction. Evaluating agent robustness in such dynamic contexts is critical for enterprise adoption of LLM-based assistants.

To address this gap, we introduce AgentChangeBench, a comprehensive evaluation framework that systematically measures how well conversational agents detect, adapt, and recover from multi-turn changes in user objectives, as well as how they tailor their instructional strategies to diverse user personas with varying levels of expertise, cooperation, and trust. Our work builds upon advances in persona-based user simulation \citep{li2016persona,zhang2018personalizing,schatzmann2007agenda} and systematic benchmark creation \citep{liang2022helm,srivastava2022bigbench}, while extending evaluation to dynamic goal shift scenarios.

Our contributions are threefold:
\begin{enumerate}
  \item \textbf{Novel evaluation focus:} We design the first benchmark explicitly testing how LLM agents handle mid-conversation goal shifts and adapt communication for diverse user personas
  \item \textbf{Comprehensive coverage:} We provide 315 systematically validated tasks across three domains (banking, retail, airline) with five personas and explicit goal shifts
  \item \textbf{Methodological framework:} We introduce evaluation protocols for goal shift recovery designed for realistic customer personas, improving the scope of multi-turn LLM assessment
  \item \textbf{Empirical study:} We run a cross-model evaluation that reveals significant divergences among state-of-the-art models in success, recovery time, efficiency, and redundancy, surfacing trade-offs that pass$^k$ alone does not capture.
\end{enumerate}

To contextualize our contributions within the existing literature and highlight the novelty of our approach, we first review related work in conversational AI evaluation, with particular focus on benchmarks that address tool use and multi-turn interactions.

\paragraph{Release.} To support reproducibility, we have released the full benchmark, evaluation harness configurations, along with all experimental artifacts as supplementary material with our paper.

\section{Related Work}

$\tau$-bench \citep{yao2024taubench} introduced simulated multi-turn interactions in retail and airline contexts, emphasizing API tool usage and providing the pass$^k$ metric for measuring consistency across runs. While effective for tool-centric evaluation, $\tau$-bench assumes static user goals and full agent control over the environment, limiting its ability to capture dynamic conversational shifts. $\tau^2$-bench \citep{barres2025tau2bench} extended this line of work by modeling telecom support scenarios requiring user-agent coordination. It introduced compositional task generation but remained restricted to a narrow set of personas and contexts, without testing adaptability to changing user goals or runtime constraints.

More open-ended benchmarks such as AgentBench \citep{liu2024agentbench} evaluate LLM-as-Agent capabilities across eight interactive environments such as operating systems, databases, and web browsing. Although it broadens domains beyond traditional customer service, AgentBench similarly evaluates agents under stable user objectives, leaving open the question of how agents behave under dynamically shifting goals or varied communication demands. Recent work has begun exploring adaptive conversation flows, but focuses primarily on single-domain interactions without the multi-domain goal-shift scenarios we address.

Taken together, these efforts provide strong foundations for tool-use and multi-turn evaluation, but they differ in how they address (or neglect) shifting goals and persona diversity. Table~\ref{tab:rw_compare} highlights this contrast using the notion of \emph{explicit goal sequences} rather than fixed goals. Each task specifies an ordered sequence of goals, the persona-conditioned user simulator enacts the corresponding shifts, and the evaluator computes Goal Shift Recovery Time from the transcript (acknowledgment, tool, outcome), reported alongside TSR, TUE, and TCRR.

\begin{table}[t]
\centering
\small
\caption{Comparison on goal dynamics, persona coverage, and tool evaluation.}
\label{tab:rw_compare}
\begin{tabular}{lccc}
\toprule
\textbf{Benchmark} & \textbf{Goal Dynamics} & \textbf{Personas} & \textbf{Tool Use} \\
\midrule
$\tau$-bench        & Static objectives      & Limited    & Domain APIs \\
$\tau^2$-bench      & Mostly static          & Several    & Domain APIs \\
AgentBench          & Task-defined (stable)  & None       & Varied tools \\
\textbf{This work}  & \textbf{Explicit goal sequences} & \textbf{Five} & Domain APIs \\
\bottomrule
\end{tabular}
\end{table}

\section{Methodology}

\subsection{Dataset Design}

We construct a benchmark of 315 curated multi-turn tasks across three domains (banking: 50 tasks, airline: 100 tasks, retail: 165 tasks) grounded in real-world customer service workflows. Each domain incorporates realistic goal transitions with five distinct user personas and explicit goal shifts. Our domain selection aligns with common customer-service workflows across financial services, retail omnichannel, and airline support.

\subsection{Task Generation}

Our benchmark construction began with hand-converted exemplars designed to capture realistic workflows, conversational turns, and domain-specific constraints. We seed many retail and airline scenarios from $\tau^2$: we reuse 50 airline and 114 retail templates, and contribute \textbf{50} newly generated scenarios (in both airline and retail). Banking coverage is entirely original (50 tasks). For the original scenarios in the airline, retail, and banking domains, tasks were produced through a combination of human-written and LLM-generated examples, guided by (1) realistic customer use cases, (2) domain-specific tools and APIs, and (3) relevant operational and policy rules. Across all three domains, we add explicit goal-sequence annotations, broaden persona coverage, and enforce uniform shift-triggering rules. We ensure goal shifts occur by setting pre-declared sequences to be executed by the user simulator, which are often explicitly signaled. 

In total, the dataset comprises \textbf{315} tasks spanning banking (\textbf{50}), airline (\textbf{100}), and retail (\textbf{165}). Each task specifies one of five personas and an explicit ordered list of goals (e.g., \texttt{["authentication","transactions","dispute"]}).

\subsection{User Personas}

To simulate realistic conversational variation, we defined five personas with distinct behavioral traits, interaction styles, and levels of cooperation. Each persona was allocated tasks proportionally, ensuring balanced coverage across the dataset. To maintain consistent persona behavior across tasks, each persona is defined by a fixed set of linguistic, attitudinal, and interaction parameters that remain stable throughout all simulations. The full set of five user personas can be found in Table \ref{tab:personas} . The distribution of personas among tasks can be found in Table \ref{tab:persona_counts}. 

\paragraph{}The uneven distribution of tasks across personas reflects the natural frequency of user archetypes encountered in enterprise support environments. In real-world service contexts, moderately experienced and efficiency-oriented users (represented by the \texttt{MEDIUM} personas) constitute the majority of interactions, whereas highly cooperative or highly resistant users occur less frequently. Consequently, \texttt{MEDIUM\_1} was assigned a larger share of tasks to ensure adequate statistical coverage of the most representative interaction type, while other personas were included to preserve behavioural diversity and capture edge-case dynamics.

\begin{table}[htbp]
  \caption{Five user personas with distinct conversational styles and task coverage.}
  \label{tab:personas}
  \centering
  \small
  \begin{tabular}{llll}
    \toprule
    \textbf{Persona} & \textbf{Characteristics} & \textbf{Interaction Style} & \\
    \midrule
    EASY\_1   & Polite, detail-oriented, step-by-step & ``Please walk me through...''       \\
    EASY\_2   & Easily distracted, casual, confused   & ``Oh wait, actually...''           \\
    MEDIUM\_1 & Business-focused, impatient, efficient& ``I need this done quickly''        \\
    MEDIUM\_2 & Curious learner, asks questions       & ``Can you teach me about...''       \\
    HARD\_1   & Suspicious, questioning, demands proof& ``How do I know this is secure?''   \\
    \bottomrule
  \end{tabular}
\end{table}

\subsection{Task Schema}

Tasks follow a declarative JSON schema specifying persona, known and unknown information, and an ordered list of goals. Each task declares a \texttt{goal\_shifts} object of the form:

\begin{verbatim}
"goal_shifts": { "required_shifts": k,
                 "goals": ["g1","g2",...,"g{k+1}"] }
\end{verbatim}

where \texttt{required\_shifts = len(goals)-1}. Transitions are triggered naturally (e.g., after four user turns on the same goal, after a helpful resolution step, or when the agent asks “anything else?”). Agents never see markers.

\paragraph{Examples.} We instantiate $>$150 unique goal labels spanning airline (\texttt{reservation}, \texttt{baggage}, \texttt{cancellation}), retail (\texttt{returns}, \texttt{exchange}, \texttt{order\_tracking}), and banking (\texttt{statements}, \texttt{fraud\_response}, \texttt{payments}). 
Tasks range from single-goal flows (\texttt{["payments"]}) to more complex sequences such as \texttt{["authentication","transactions","dispute"]} or \texttt{["insights","fraud\_response"]}.

\paragraph{User/agent control.}
Across banking, retail, and airline, the user never issues tool calls. They only disclose facts already present in the task’s \texttt{known\_info} (e.g., name, phone, order/booking IDs), while the assistant performs all tool interactions (e.g., \texttt{unlock\_card}, \texttt{return\_delivered\_order\_items}, \texttt{update\_reservation\_flights}).

\subsection{Evaluation Harness}

We employed the $\tau^2$-bench evaluation harness as the backbone of our experimental setup. The harness provided a controlled environment for executing our tasks, enforcing constraints such as one-tool-per-turn, policy adherence, and correct sequencing of goal shifts. This allowed us to systematically test how agents re-plan when confronted with mid-dialogue goal changes and whether they adjust communication strategies to match user personas.

\section{Results and Evaluation Metrics}

\subsection{Motivation and Problem Statement}

While existing benchmarks like $\tau^2$-bench provide valuable insights into conversational agent reliability, they rely on binary success metrics that fail to capture the nuanced performance characteristics critical for enterprise deployment. The pass$^k$ metric, while useful for consistency assessment, treats all failures equally, whether an agent makes substantial progress but fails on minor details or completely misunderstands the task. This limitation becomes particularly apparent in dynamic, multi-turn conversations where agents must adapt to goal shifts, use tools efficiently, and maintain communication quality.

Our evaluation framework addresses these limitations by introducing multi-dimensional metrics that capture not only task completion but also efficiency, redundancy, and robustness under goal shifts. This enables more nuanced analysis of agent capabilities with actionable insights for deployment.

\subsection{Key Contributions}

Our evaluation framework makes four key contributions:

\textbf{1. Multi-Channel Success Assessment.} We replace binary success metrics with a weighted average across three evaluation channels: communication quality (25\%), action execution (45\%), and behavioral compliance (30\%). This provides partial credit for substantial progress while maintaining sensitivity to performance variations. Communication quality is assessed by a LLM-as-a-judge-approach, where a judge model (gpt-4o-mini) reviews the full transcript of user and agent turns and scores them for relevance, clarity, and helpfulness in terms of adherence to user intent.

\textbf{2. Tool Usage Efficiency Metrics.} We introduce TUE and TCRR metrics that measure how effectively agents leverage available tools. Unlike traditional benchmarks that focus solely on success, our metrics capture tool selection accuracy, parameter validity, and redundancy patterns critical for cost control and responsiveness. This approach aligns with recent work on tool-augmented LLM evaluation (Cooper et al., 2023) and alignment scoring for conversational agents (Williams \& Thompson, 2023).

\textbf{3. Goal Shift Recovery Assessment.} We develop the GSRT metric that measures recovery time across three dimensions: acknowledgment, tool usage, and goal achievement. This addresses a critical gap in existing benchmarks by quantifying adaptation latency under dynamic goal changes.

\textbf{4. Enterprise Deployment Alignment.} Our metrics system is specifically designed to align with real-world deployment requirements, emphasizing efficiency, robustness, and communication quality over theoretical capabilities.

\subsection{Evaluation Metrics}

We adopt a multi-dimensional evaluation framework that captures not only task completion but also efficiency, redundancy, and robustness under goal shifts. This extends prior work such as $\tau^2$-bench, which primarily relies on pass$^k$ success rates, by introducing more realistic metrics aligned with enterprise deployment needs.

\begin{table}[htbp]
\centering
\small
\caption{Comparison of evaluation metrics between $\tau^2$-bench and AgentChangeBench.}
\begin{tabular}{|p{2.5cm}|p{4.5cm}|p{4.5cm}|}
\hline
\textbf{Dimension} & \textbf{$\tau^2$-bench Metrics} & \textbf{AgentChangeBench Metrics} \\
\hline
Task completion & pass$^k$: fraction of $k$ runs that succeed & TSR: weighted average of three evaluation channels \\
\hline
Efficiency & Not measured explicitly & TUE: tool correctness and parameter validity \\
\hline
Redundancy & Not measured explicitly & TCRR: fraction of duplicate tool calls within 3-turn window \\
\hline
Adaptation to goal shifts & Only implicit in success/failure outcomes & GSRT: recovery time across acknowledgment, tool usage, and outcome \\
\hline
Robustness analysis & Ablations across modes (No-User vs Default) & Retention/drop in TSR and TUE across clean vs shifted conditions \\
\hline
\end{tabular}
\end{table}

\textbf{Task Success Rate (TSR).} TSR measures whether the agent completes the intended task across three evaluation channels. For each simulation, we compute a weighted average:
\begin{equation}
\text{TSR} = 0.25 \times \text{communicate\_info\_rate} + 0.45 \times \text{action\_rate} + 0.30 \times \text{nl\_assertion\_rate}
\end{equation}
 We overweight \emph{action} (0.45) because correct tool execution drives environment state changes and downstream cost/risk, and we give \emph{nl\_assertion} (0.30) substantial mass to reflect behavioral/policy compliance. We intentionally downweight \emph{communicate\_info} (0.25): in our setting it is largely a check that the agent surfaces values already returned by tools; overemphasizing it can encourage verbose echoing without improving outcomes.

\textbf{Tool Usage Efficiency (TUE).}
We report the two subcomponents explicitly: (i) \emph{tool correctness} $T$, the fraction of tool calls that execute successfully, and (ii) \emph{parameter validity} $P$, the fraction of calls whose argument vectors satisfy the schema (required fields present, types/ranges valid).
We also provide a composite score
\begin{equation}
\mathrm{TUE} \;=\; 0.6\,T \;+\; 0.4\,P.
\end{equation}
Because $P$ is near-saturated in our data, the composite primarily tracks $T$. Across 315 tasks, mean $P{=}0.986$ with $98.6\%$ of traces at or above $0.95$ (ceiling). In contrast, mean $T{=}0.952$ with a long tail: $4.3\%$ of traces fall below $0.70$ and $14/15$ runs contain at least one such low-$T$ case. Hence, we \emph{separately} report $T$ and $P$, using TUE as a summary.

\textbf{Tool-Call Redundancy Ratio (TCRR).} Redundant actions are a common failure mode in multi-turn dialogues, leading to wasted cost, longer conversations, and user frustration. TCRR measures the fraction of tool calls that are exact duplicates within a 3-turn window or exceed the batch threshold of 2 calls to the same function. This explicitly penalizes inefficiency and incoherent state management, surfacing behaviors that would remain hidden if evaluation focused only on success or failure.

\textbf{Goal-Shift Recovery Turns (GSRT).} GSRT measures recovery time after a shift along three axes. Let the $s$-th user-initiated shift occur at absolute turn $\tau_s$ (the user utterance that introduces goal $g_{s+1}$). Define first-hitting times over subsequent \emph{agent} turns:
\[
\mathrm{ack}_s \;=\; \min\{\, t>\tau_s \mid \text{agent explicitly acknowledges/targets } g_{s+1}\,\} - \tau_s,
\]
\[
\mathrm{tool}_s \;=\; \min\{\, t>\tau_s \mid \text{agent calls a tool relevant to } g_{s+1}\,\} - \tau_s,
\]
\[
\mathrm{outcome}_s \;=\; \min\{\, t>\tau_s \mid \text{evaluator marks } g_{s+1}\text{ achieved}\,\} - \tau_s,
\]
with any missing event set to $\infty$. A shift is counted as \emph{recovered} iff acknowledgment occurs ($\mathrm{ack}_s < \infty$) and no transfer-to-human occurs. The \emph{recovery rate} is the fraction of recovered shifts. Tool usage and outcome achievement are measured separately but do not affect recovery success classification.

\emph{Worked example.} If the user shifts at turn $\tau{=}10$ and the agent acknowledges at 12, makes the first relevant tool call at 13, and achieves the outcome at 15, then $(\mathrm{ack},\mathrm{tool},\mathrm{outcome})=(2,3,5)$. This shift would be counted as recovered since acknowledgment occurred and no transfer was attempted.

\paragraph{Failure modes.}
Our metrics surface three recurrent breakdowns:

\emph{Late shift detection.} On airline-new, Gemini recovers only 48.6\% of goal shifts (vs.\ 92.2\% for GPT-4o), often persisting with the prior plan for multiple turns.

\emph{Redundant tool calls.} Retail-new shows extreme TCRR: 89.1\% (GPT-4o) and 66.5\% (Gemini), driven by repeated identical lookups across adjacent turns.

\emph{Over-confirmations.} Communication subscore collapses on retail-new for GPT-4o (11.44\%), where excessive confirmation prompts crowd out required information delivery despite high action accuracy.

\begin{itemize}
\item \textbf{Late shift detection:} User: ``Actually, can we do something else first?''  
Agent (2 turns later): keeps listing transactions $\Rightarrow$ GSRT\_ack=2.
\item \textbf{Redundant calls:} Agent issues \texttt{get\_transactions(acc\_001)} on turns $t$ and $t\!+\!1$ with identical params $\Rightarrow$ TCRR increases.
\item \textbf{Over-confirmations:} Three consecutive ``Please confirm'' prompts after a completed step $\Rightarrow$ lower communication channel in TSR.
\end{itemize}

In $\tau^2$-bench, reliability is measured with pass$^k$, the fraction of $k$ independent runs that succeed. While useful for consistency, pass$^k$ collapses all recovery paths into a single score, treating an agent that meanders for ten turns the same as one that adapts immediately. GSRT fills this gap by measuring adaptation latency across multiple recovery stages, enabling finer comparisons of resilience and responsiveness among models with similar pass$^k$ scores but different levels of conversational robustness.

\textbf{Extended Robustness Metrics.} To further assess adaptability, we report additional per-action and aggregate measures:

\begin{itemize}
    \item \textbf{Per-action scores}: correct tool call score (CTCS), parameter accuracy score (PAS), and variants of TUE (first-turn, all-turn)
    \item \textbf{Retention/drop analysis}: TSR and TUE are compared across clean vs. shifted conditions, with retention and drop values quantifying robustness to user goal changes
    \item \textbf{Aggregates}: macro-averages of TUE, GSRT, and retention/drop statistics across tasks and personas
\end{itemize}

\subsection{Results and Analysis}

\subsubsection{Results Across Model Families}

We evaluate our metrics framework across multiple language models to demonstrate its effectiveness in distinguishing between different agent capabilities. Our evaluation covers 315 tasks across three domains (banking, retail, airline) with explicit goal shifts and comprehensive metrics. Table \ref{tab:efficiency_full} presents results from our large-scale simulation study.

\begin{table}[t]
\centering
\small
\caption{Overall performance (TSR) across domains by model; values averaged over old+new where applicable. Claude-3.7-Sonnet is best in all domains with the largest margin in Retail (\textbf{79.57}\%). GPT-4o is consistently second. Gemini-2.5-Flash is weakest in Banking (47.36\%).}
\label{tab:overall_tsr}
\begin{tabular}{lccc}
\toprule
\textbf{Domain} & \textbf{GPT-4o} & \textbf{Claude-3.7-Sonnet} & \textbf{Gemini-2.5-Flash} \\
\midrule
Banking & 51.25\% & 57.54\% & 47.36\% \\
Airline & 62.19\% & 65.14\% & 46.98\% \\
Retail  & 56.48\% & \textbf{79.57\%} & 58.03\% \\
\bottomrule
\end{tabular}
\end{table}

\begin{table}[t]
\centering
\small
\caption{Goal-shift sensitivity on new tasks only. Recovery is GSRT-based shift recovery rate. TCRR is duplicate tool calls (lower is better). Best Airline recovery: GPT-4o \textbf{92.2}\% with low TCRR 13.54\%. Best Retail TSR: Claude-3.7-Sonnet \textbf{79.57}\% with recovery 89.5\%. Retail redundancy is high for GPT-4o 89.14\% and Gemini-2.5-Flash 66.45\%.}
\label{tab:gs_new_only}
\begin{tabular}{lcccc}
\toprule
\textbf{Domain (new)} & \textbf{Model} & \textbf{TSR} & \textbf{Recovery} & \textbf{TCRR} \\
\midrule
Airline & GPT-4o             & 59.53\% & \textbf{92.2\%} & 13.54\% \\
Airline & Claude-3.7-Sonnet  & 69.90\% & 79.2\% & 24.11\% \\
Airline & Gemini-2.5-Flash   & 39.97\% & 48.6\% & 14.46\% \\
\midrule
Retail  & GPT-4o             & 50.68\% & 88.0\% & 89.14\% \\
Retail  & Claude-3.7-Sonnet  & \textbf{79.57\%} & 89.5\% & 65.38\% \\
Retail  & Gemini-2.5-Flash   & 51.26\% & 53.5\% & 66.45\% \\
\bottomrule
\end{tabular}
\end{table}

\textbf{Note:} Results based on 315 comprehensive tasks (banking: 50, airline: 100, retail: 165) across three domains and three major LLM families. TSR = Task Success Rate, GSRT = Goal Shift Recovery Time (turns), Component scores range from 0-1.

\textbf{Key findings.}
(1) \emph{New goal-shifted tasks are harder:} pass$^k$ frequently drops to 0.0 on new sets (e.g., airline/retail for GPT-4o and Gemini), yet TSR remains 40--60\%, highlighting partial-progress that pass$^k$ obscures.
(2) \emph{Recovery matters:} recovery rates span 48.6\% (airline, Gemini) to 92.2\% (airline, GPT-4o) on new tasks, explaining large gaps among models with similar TSR.
(3) \emph{Redundancy is domain-skewed:} TCRR is low in airline new (13.5\% GPT-4o; 17.6\% Gemini) but very high in retail new (66--89\%), revealing inefficient repeated tool use even when TUE is saturated.
(4) \emph{Parameter accuracy saturates:} parameter validity is 100\% across runs; TUE differences largely reflect tool correctness. This ceiling effect motivates separate redundancy tracking (TCRR). With PA effectively at ceiling (mean $0.986$, $98.6\%\ge 0.95$), observed differences in TUE are driven by TC; the across-task TC box in Fig.~\ref{fig:tue_box} exposes long tails that a single averaged TUE score would otherwise hide.

\subsubsection{Domain-Specific Performance Analysis}

\textbf{Airline.} Claude and GPT-4o lead on TSR (69.9\% and 59.5\% on new; 60.4\% and 64.8\% on old), with \emph{fast adaptation} on goal shifts (recovery 79.2--92.2\%). Airline shows the lowest redundancy (TCRR 13.5--24.1\% on new).

\textbf{Retail.} Claude is strongest (79.6\% TSR new; 79.6\% old) with high recovery (89.5\%). GPT-4o and Gemini achieve mid-50s TSR on new but exhibit very high redundancy (TCRR 89.1\% and 66.5\%), indicating costly repeated calls.

\textbf{Banking.} Banking remains hardest (26.9--57.5\% TSR). Recovery is moderate when measured (e.g., 58.5\% Claude; 79.3\% GPT-4o), and redundancy is high (TCRR 61.5\% GPT-4o; 71.8\% Claude), reflecting complex, multi-step flows.

\subsubsection{Dataset Quality Analysis}

We conduct comprehensive analysis of our dataset quality compared to existing benchmarks, demonstrating the enhanced coverage and evaluation capabilities of AgentChangeBench.

\textbf{Task Coverage and Diversity.} Our dataset comprises 315 tasks across three domains, significantly expanding upon $\tau$-bench's 234 tasks and $\tau^2$-bench's 105 tasks. Table \ref{tab:dataset_comparison} provides detailed comparison.

\begin{table}[htbp]
\centering
\footnotesize
\caption{Dataset comparison across conversational AI benchmarks.}
\begin{tabular}{lccccc}
\toprule
\textbf{Benchmark} & \textbf{Tasks} & \textbf{Domains} & \textbf{Personas} & \textbf{Goal Shifts} & \textbf{Metrics} \\
\midrule
$\tau$-bench & 234 & 2 & 3 & None & pass$^k$ only \\
$\tau^2$-bench & 105 & 1 & 5 & Implicit & pass$^k$ + modes \\
AgentChangeBench & 315 & 3 & 5 & Explicit & Multi-dimensional \\
\bottomrule
\end{tabular}
\label{tab:dataset_comparison}
\end{table}

\textbf{Enhanced Evaluation Granularity.} We report aggregate metrics in Tables~\ref{tab:overall_tsr} and~\ref{tab:gs_new_only}.

\textbf{Goal Shift Recovery Analysis.} Goal-shift sensitivity on new tasks (recovery and redundancy) is summarized in Table~\ref{tab:gs_new_only}.

\begin{table}[ht]
\centering
\small
\caption{Performance across user personas. TUE is high across personas; GSRT Recovery rates are computed from goal-shifted runs.}
\label{tab:persona_analysis}
\begin{tabular}{lccc}
\toprule
\textbf{Persona} & \textbf{TSR} & \textbf{TUE} & \textbf{GSRT Recovery Rate} \\
\midrule
\texttt{EASY\_1}   & 0.533 & 0.960 & 0.849 \\
\texttt{EASY\_2}   & 0.475 & 0.971 & 0.585 \\
\texttt{MEDIUM\_1} & 0.554 & 0.978 & 0.916 \\
\texttt{MEDIUM\_2} & 0.580 & 0.990 & 0.756 \\
\texttt{HARD\_1}   & 0.430 & 0.946 & 0.585 \\
\bottomrule
\end{tabular}
\end{table}

\textbf{Persona-Based Analysis.} Different user personas exhibit distinct interaction patterns that our metrics capture. Table \ref{tab:persona_analysis} shows performance across the five personas (EASY\_1, EASY\_2, MEDIUM\_1, MEDIUM\_2, HARD\_1).

MEDIUM\_2 has the highest TSR (0.580) and the strongest recovery (0.922), followed by MEDIUM\_1 (TSR 0.554, recovery 0.916). EASY\_1 and EASY\_2 are mid-pack on TSR (0.533 and 0.475) with solid recovery (0.792 and 0.907). HARD\_1 shows the lowest TSR (0.430) and lower recovery (0.793). TUE is uniformly high across personas (0.946–0.990). Average turns range from 19.2 (HARD\_1) to 28.7 (EASY\_1).

\textbf{Task Distribution and Balance.} Our dataset maintains comprehensive representation across domains with 315 total tasks. Banking domain includes 50 tasks with focused evaluation, focusing on authentication, transactions, payments, and fraud response scenarios. Airline domain contains 100 tasks evaluated across Claude-3.7-Sonnet, Gemini Flash 2.5, and GPT-4o models, covering booking, reservations, modifications, and support workflows. Retail domain comprises 165 tasks evaluated with GPT-4o, encompassing order management, returns, exchanges, and customer support scenarios. This large-scale evaluation provides robust statistical evidence of model capabilities across diverse enterprise service contexts.

\subsection{Comparative Analysis and Insights}

Our evaluation framework reveals performance characteristics that traditional binary metrics miss. For instance, agents with similar pass$^k$ scores can exhibit dramatically different TUE and TCRR values, indicating varying levels of operational efficiency. Similarly, GSRT analysis shows that some agents achieve similar final success rates but require significantly different recovery times under goal shifts.

This granular analysis enables more informed deployment decisions. Organizations can optimize agents for specific scenarios: financial services companies might prioritize TUE and TCRR for cost control, while customer service organizations might emphasize GSRT and communication quality for user experience.

\textbf{Summary.} Together, these metrics evaluate four complementary dimensions of performance necessary for agent deployment in dynamic, enterprise-grade conversational settings: (1) Can the agent succeed? (TSR), (2) How efficiently does it use tools? (TUE), (3) Does it avoid waste? (TCRR), and (4) Can it adapt quickly under evolving user goals? (GSRT and retention/drop analysis). By combining efficiency, redundancy, and recovery time across multiple dimensions, our framework advances beyond prior benchmarks, offering a more realistic and actionable view of agent performance in dynamic multi-turn conversations.

Having demonstrated the effectiveness and comprehensiveness of our evaluation framework through extensive experimentation and analysis, we now summarize our contributions and discuss their implications for the future of conversational AI evaluation.

\section{Limitations and Future Work}
\label{sec:limitations}

\textbf{Persona difficulty and coverage.}
Our five personas vary tone and cooperation, but they are still relatively benign. They do not yet stress adversarial, deceptive, hostile, or policy-pushing behaviors, and they rarely force long-horizon memory or multi-goal juggling. We plan to add \emph{hard} personas (e.g., adversarial or non-cooperative users, conflicting instructions, frequent interruptions, implicit constraints, multilingual switches) to better probe boundary cases and safety. Persona coverage across tasks can be evenly distributed in future research as an additional improvement.

\textbf{Domain and tool scope.}
AgentChangeBench currently focuses on customer-service style workflows (banking, retail, airline) with domain APIs. We do \emph{not} include other important tool classes such as IDE/code-editor actions, OS/shell control, spreadsheet/BI tools, browsers, or robotics/IoT controllers, and the harness does not yet use a unified tool protocol (e.g., MCP). Future releases will broaden coverage to these tool types and provide MCP-compatible adapters so agents can operate across heterogeneous tools with a single interface.

\textbf{Goal-shift specification.}
Goal shifts are pre-declared sequences executed by the user simulator and are often explicitly signaled. We do not yet evaluate detection of \emph{implicit} goal drift, overlapping/interleaved objectives, or conflicts between goals. We will introduce latent and ambiguous shifts, partial reversions, and concurrent subgoals to test plan repair under uncertainty.

\textbf{Model and coverage breadth.}
We evaluate three major model families on three domains. Expanding to more model sizes and architectures (including open-weight models) and to additional domains (e.g., healthcare, education, technical support) will improve generality.

\textbf{Summary.}
Despite these constraints, AgentChangeBench surfaces adaptation and efficiency gaps that success-only metrics miss. Broadening personas, tools (including code/OS tools), protocol support (MCP), and evaluation settings is a direct path to harder, more realistic benchmarks.


\section{Conclusion}

We introduced AgentChangeBench, a benchmark for evaluating conversational agents under dynamic goal shifts. Our 315 tasks span banking, retail, and airline domains with five distinct personas, each annotated with explicit goal sequences. Beyond binary success, we propose four complementary metrics (TSR, TUE, TCRR, and GSRT) that capture success, efficiency, redundancy, and recovery. 

Experiments across three major LLM families highlight clear differences in robustness and adaptation: Claude-3.7-Sonnet recovers fastest, GPT-4o delivers balanced cross-domain performance, and Gemini-2.5-Flash lags in banking but remains competitive in retail. These results demonstrate the need for multi-dimensional evaluation to surface tradeoffs that pass$^k$ alone cannot reveal. 

Future work can extend AgentChangeBench to new domains as well as develop methods for automated task generation. Another promising direction would be incorporating multilingual settings, to better capture the challenges of human–AI interaction in realistic settings.

\begin{ack}
We thank the $\tau$-bench and $\tau^2$-bench teams for releasing their tasks and evaluation harness, which served as the foundation for our extensions with explicit goal shifts and persona coverage. We also acknowledge the MultiWOZ community for prior benchmarks that informed our design choices.
\end{ack}

\nocite{barres2025tau2bench,liu2024agentbench,yao2024taubench,liang2022helm,
srivastava2022bigbench,ffiec2021auth,eric2020multiwoz21,zhang2018personalizing,
li2016persona,schatzmann2007agenda,yao2023react,schick2023toolformer,
zhou2023webarena,shridhar2020alfworld,hendrycks2021mmlu,wei2022chainofthought,
shinn2023reflexion,shen2023hugginggpt,patil2023gorilla,devlin2019bert}

\nocite{li2023apibank,deng2023mind2web,yao2022webshop,zhao2023browsergym,
luo2024osworld,zhou2023appagent,press2022selfask,wu2023autogen,shen2023taskmatrix,gao2022pal}



\bibliographystyle{unsrtnat}
\bibliography{refs}

\begin{thebibliography}{30}
\providecommand{\natexlab}[1]{#1}
\providecommand{\url}[1]{\texttt{#1}}
\expandafter\ifx\csname urlstyle\endcsname\relax
  \providecommand{\doi}[1]{doi: #1}\else
  \providecommand{\doi}{doi: \begingroup \urlstyle{rm}\Url}\fi

\bibitem[Li et~al.(2016)Li, Galley, Brockett, Spithourakis, Gao, and Dolan]{li2016persona}
Jiwei Li, Michel Galley, Chris Brockett, Georgios Spithourakis, Jianfeng Gao, and Bill Dolan.
\newblock A persona-based neural conversation model.
\newblock In \emph{Proceedings of the 54th Annual Meeting of the Association for Computational Linguistics (ACL)}, pages 994--1003, 2016.

\bibitem[Zhang et~al.(2018)Zhang, Dinan, Urbanek, Szlam, Kiela, and Weston]{zhang2018personalizing}
Saizheng Zhang, Emily Dinan, Jack Urbanek, Arthur Szlam, Douwe Kiela, and Jason Weston.
\newblock Personalizing dialogue agents: I have a dog, do you have pets too?
\newblock In \emph{Proceedings of the 56th Annual Meeting of the Association for Computational Linguistics (ACL)}, pages 2204--2218, 2018.

\bibitem[Schatzmann et~al.(2007)Schatzmann, Thomson, and Young]{schatzmann2007agenda}
Jost Schatzmann, Blaise Thomson, and Steve Young.
\newblock Agenda-based user simulation for bootstrapping a pomdp dialogue system.
\newblock In \emph{Proceedings of the 8th SIGdial Workshop on Discourse and Dialogue}, pages 149--152, 2007.

\bibitem[Liang et~al.(2022)Liang, Bommasani, Zha, Newman, et~al.]{liang2022helm}
Percy Liang, Rishi Bommasani, Sheng Zha, Benjamin Newman, et~al.
\newblock Holistic evaluation of language models.
\newblock \emph{arXiv preprint arXiv:2211.09110}, 2022.

\bibitem[Srivastava et~al.(2022)Srivastava, Rastogi, Rao, et~al.]{srivastava2022bigbench}
Aarohi Srivastava, Abhinav Rastogi, Abhinav Rao, et~al.
\newblock Beyond the imitation game: Quantifying and extrapolating the capabilities of language models.
\newblock \emph{arXiv preprint arXiv:2206.04615}, 2022.

\bibitem[Yao et~al.(2024)Yao, Shinn, Razavi, and Narasimhan]{yao2024taubench}
Shunyu Yao, Noah Shinn, Pedram Razavi, and Karthik Narasimhan.
\newblock $\tau$-bench: A benchmark for tool-agent-user interaction in real-world domains.
\newblock \emph{arXiv preprint arXiv:2406.12045}, 2024.

\bibitem[Barres et~al.(2025)Barres, Dong, Ray, Si, and Narasimhan]{barres2025tau2bench}
Victor Barres, Honghua Dong, Soham Ray, Xujie Si, and Karthik Narasimhan.
\newblock $\tau^2$-bench: Evaluating conversational agents in a dual-control environment.
\newblock \emph{arXiv preprint arXiv:2506.07982}, 2025.

\bibitem[Liu et~al.(2024)Liu, Yu, Zhang, Yuan, Zhao, Sun, and Tang]{liu2024agentbench}
Xiao Liu, Haotian Yu, Hao Zhang, Yeyun Yuan, Wayne Zhao, Ming Sun, and Jie Tang.
\newblock Agentbench: Evaluating llms as agents.
\newblock In \emph{Proceedings of the International Conference on Learning Representations (ICLR 2024)}, 2024.

\bibitem[{Federal Financial Institutions Examination Council}(2021)]{ffiec2021auth}
{Federal Financial Institutions Examination Council}.
\newblock Authentication and access to financial institution services and systems.
\newblock Technical report, FFIEC, August 2021.
\newblock Guidance document.

\bibitem[Eric et~al.(2020)Eric, Goel, Paul, Kumar, Sethi, Agarwal, Gao, and Hakkani-T{\"u}r]{eric2020multiwoz21}
Mihail Eric, Rahul Goel, Shachi Paul, Abhishek Kumar, Anuj Sethi, Sanchit Agarwal, Shuyang Gao, and Dilek Hakkani-T{\"u}r.
\newblock Multiwoz 2.1: A consolidated multi-domain dialogue dataset with state corrections and state tracking baselines.
\newblock In \emph{Proceedings of the 12th International Conference on Language Resources and Evaluation (LREC 2020)}, pages 422--428, 2020.

\bibitem[Yao et~al.(2023)Yao, Zhao, Yu, Du, Shafran, Narasimhan, and Cao]{yao2023react}
Shunyu Yao, Jeffrey Zhao, Dian Yu, Nan Du, Izhak Shafran, Karthik Narasimhan, and Yuan Cao.
\newblock React: Synergizing reasoning and acting in language models.
\newblock \emph{arXiv preprint arXiv:2210.03629}, 2023.

\bibitem[Schick et~al.(2023)Schick, Dwivedi-Yu, Schulz, Rae, Lewis, Peters, et~al.]{schick2023toolformer}
Timo Schick, Jane Dwivedi-Yu, Samuel Schulz, Jack Rae, Mike Lewis, Matthew Peters, et~al.
\newblock Toolformer: Language models can teach themselves to use tools.
\newblock \emph{arXiv preprint arXiv:2302.04761}, 2023.

\bibitem[Zhou et~al.(2023{\natexlab{a}})Zhou, Deng, Zhang, Cui, et~al.]{zhou2023webarena}
Xuhui Zhou, Jiawei Deng, Ruiqi Zhang, Yuwei Cui, et~al.
\newblock Webarena: A realistic web environment for building autonomous agents.
\newblock \emph{arXiv preprint arXiv:2307.13854}, 2023{\natexlab{a}}.

\bibitem[Shridhar et~al.(2020)Shridhar, Yuan, Cote, Bisk, et~al.]{shridhar2020alfworld}
Mohit Shridhar, Xinlei Yuan, Marc-Alexandre Cote, Yonatan Bisk, et~al.
\newblock Alfworld: Aligning text and embodied environments for interactive learning.
\newblock \emph{arXiv preprint arXiv:2010.03768}, 2020.

\bibitem[Hendrycks et~al.(2021)Hendrycks, Burns, Basart, et~al.]{hendrycks2021mmlu}
Dan Hendrycks, Collin Burns, Steven Basart, et~al.
\newblock Measuring massive multitask language understanding.
\newblock \emph{arXiv preprint arXiv:2009.03300}, 2021.

\bibitem[Wei et~al.(2022)Wei, Wang, Schuurmans, et~al.]{wei2022chainofthought}
Jason Wei, Xuezhi Wang, Dale Schuurmans, et~al.
\newblock Chain-of-thought prompting elicits reasoning in large language models.
\newblock \emph{arXiv preprint arXiv:2201.11903}, 2022.

\bibitem[Shinn et~al.(2023)Shinn, Labash, Gopinath, et~al.]{shinn2023reflexion}
Noa Shinn, Zachary Labash, Yewen Gopinath, et~al.
\newblock Reflexion: An autonomous agent with dynamic memory and self-reflection.
\newblock \emph{arXiv preprint arXiv:2303.11366}, 2023.

\bibitem[Shen et~al.(2023{\natexlab{a}})Shen, Song, Ma, et~al.]{shen2023hugginggpt}
Yongliang Shen, Kaitao Song, Xu~Ma, et~al.
\newblock Hugginggpt: Solving ai tasks with chatgpt and its friends in huggingface.
\newblock \emph{arXiv preprint arXiv:2303.17580}, 2023{\natexlab{a}}.

\bibitem[Patil et~al.(2023)Patil, Zhang, Gong, et~al.]{patil2023gorilla}
Shishir~G. Patil, Shubham Zhang, Koushik Gong, et~al.
\newblock Gorilla: Large language model connected with massive apis.
\newblock \emph{arXiv preprint arXiv:2305.15334}, 2023.

\bibitem[Devlin et~al.(2019)Devlin, Chang, Lee, and Toutanova]{devlin2019bert}
Jacob Devlin, Ming-Wei Chang, Kenton Lee, and Kristina Toutanova.
\newblock Bert: Pre-training of deep bidirectional transformers for language understanding.
\newblock In \emph{Proceedings of NAACL-HLT}, 2019.

\bibitem[Li et~al.(2023)Li, Liu, Zhang, Tang, et~al.]{li2023apibank}
Yujia Li, Yongchao Liu, Hao Zhang, Jie Tang, et~al.
\newblock Api-bank: Benchmarking plug-and-play tool-use for large language models.
\newblock \emph{arXiv preprint arXiv:2304.08244}, 2023.

\bibitem[Deng et~al.(2023)Deng, Zhang, Chen, et~al.]{deng2023mind2web}
Jiaqi Deng, Shichao Zhang, Xinlu Chen, et~al.
\newblock Mind2web: Towards a generalist agent for the web.
\newblock \emph{arXiv preprint arXiv:2306.06070}, 2023.

\bibitem[Yao et~al.(2022)Yao, Zhao, Yu, Narasimhan, and Cao]{yao2022webshop}
Shunyu Yao, Jeffrey Zhao, Dian Yu, Karthik Narasimhan, and Yuan Cao.
\newblock Webshop: Towards scalable real-world web interaction with language models.
\newblock In \emph{Proceedings of the International Conference on Learning Representations (ICLR)}, 2022.

\bibitem[Zhao et~al.(2023)Zhao, Liu, Yu, et~al.]{zhao2023browsergym}
Wayne Zhao, Xiao Liu, Haotian Yu, et~al.
\newblock Browsergym: A toolkit for reproducible web agent benchmarking.
\newblock \emph{arXiv preprint arXiv:2309.13014}, 2023.

\bibitem[Luo et~al.(2024)Luo, Wang, Li, et~al.]{luo2024osworld}
Sheng Luo, Yuntao Wang, Jialin Li, et~al.
\newblock Osworld: Benchmarking open-world computer agents.
\newblock \emph{arXiv preprint arXiv:2404.07972}, 2024.

\bibitem[Zhou et~al.(2023{\natexlab{b}})Zhou, Li, Wang, et~al.]{zhou2023appagent}
Yujin Zhou, Chao Li, Wen Wang, et~al.
\newblock Appagent: Multimodal agents for operating mobile apps.
\newblock \emph{arXiv preprint arXiv:2308.00676}, 2023{\natexlab{b}}.

\bibitem[Press et~al.(2022)Press, Zhang, Min, et~al.]{press2022selfask}
Ofir Press, Muru Zhang, Sewon Min, et~al.
\newblock Measuring and narrowing the compositionality gap in language models with self-ask.
\newblock \emph{arXiv preprint arXiv:2210.03350}, 2022.

\bibitem[Wu et~al.(2023)Wu, Song, Jiang, et~al.]{wu2023autogen}
Tianyi Wu, Haisen Song, Zihan Jiang, et~al.
\newblock Autogen: Enabling next-generation llm applications via multi-agent conversation.
\newblock \emph{arXiv preprint arXiv:2308.08155}, 2023.

\bibitem[Shen et~al.(2023{\natexlab{b}})Shen, Chen, Liu, et~al.]{shen2023taskmatrix}
Yujin Shen, Yichong Chen, Xueguang Liu, et~al.
\newblock Taskmatrix.ai: Bridging models and millions of apis.
\newblock \emph{arXiv preprint arXiv:2308.07702}, 2023{\natexlab{b}}.

\bibitem[Gao et~al.(2022)Gao, Madaan, Zhou, et~al.]{gao2022pal}
Luyu Gao, Aman Madaan, Shuyan Zhou, et~al.
\newblock Pal: Program-aided language models.
\newblock In \emph{Proceedings of the International Conference on Machine Learning (ICML)}, 2022.

\end{thebibliography}

\newpage

\appendix
\section*{Appendix}

\section{Dataset and Task Details}

\subsection{Task Generation Methodology}

Our task generation process follows a systematic approach to ensure comprehensive coverage across domains and personas. Each domain undergoes a five-stage development process:

\textbf{Stage 1: Domain Analysis.} We analyze real-world customer service scenarios to identify common user intents, required tools, and typical conversation flows. This analysis forms the foundation for task design.

\textbf{Stage 2: Tool Definition.} Based on domain analysis, we define a comprehensive set of tools that agents can use to accomplish user goals. Tools are designed to reflect real-world APIs and capabilities.

\textbf{Stage 3: Task Template Creation.} We create task templates that combine user scenarios, required tools, and evaluation criteria. Each template specifies the initial state, user goal, required actions, and success conditions.

\textbf{Stage 4: Persona Integration.} Tasks are enhanced with persona-specific variations that reflect different user characteristics, technical expertise levels, and interaction patterns.

\textbf{Stage 5: Goal Shift Integration.} We systematically introduce goal shifts with realistic adaptation scenarios, ensuring challenging but achievable goal transitions.

\subsubsection{Detailed Task Generation Process}

Beyond manual authoring, we also generate new tasks using an LLM-assisted pipeline. 
We provide the model with the JSON schema that defines task structure (persona, goals, action sets, known and unknown information, and evaluation fields), 
along with a set of five Markdown reference files describing how each evaluation metric operates and what tools are available within the domain. 
The model is instructed to generate ten candidate tasks per batch under these constraints. 

Each batch then undergoes manual review to ensure quality, coherence, and reproducibility. 
Specifically, we verify that (1) every tool in an action set includes valid parameters and produces consistent data with the domain database, 
(2) the user agent’s \texttt{known\_info} contains all information the simulated user is expected to disclose, and 
(3) the \texttt{communicate\_info} field contains only factual, numerical, or enum-like outputs derived from tool responses rather than items already present in \texttt{known\_info} 
(e.g., customer IDs, booking references, transaction amounts). 
This prevents score inflation by ensuring that \texttt{communicate\_info} metrics are based solely on information revealed through successful tool use, 
not pre-existing user knowledge. 
This hybrid generation-and-review workflow ensures all tasks are syntactically valid, semantically coherent, and fully executable within the evaluation harness.

\subsection{Example task}

\begin{lstlisting}[style=markdown,xleftmargin=2em]
Task ID: 10_banking_cards_medium_1_dispute_001

Description:
- Purpose: Card unlock request to dispute filing - MEDIUM_1 persona (business-focused)
- Relevant Policies: Security protocols before unlock; dispute handling

User Scenario:
- Persona: MEDIUM_1
- Domain: banking
- Reason for Call: Unlock card, then file dispute for unauthorized charge

Known Information:
- Name: Taylor Johnson
- Phone: +15551230987
- Date of Birth: 1991-05-06
- Email: user.003@example.com
- Unauthorized Transaction: $149.99 at 'SUSPICIOUS MERCHANT 123' on 2025-06-18 at 16:25 (Transaction ID: tx_303)

Unknown Information:
- Dispute process details and resolution timeline

Task Instructions:
1. Request to unlock your card
2. File a dispute for the unauthorized transaction (tx_303) for $149.99 at 'SUSPICIOUS MERCHANT 123' from 2025-06-18

Goal Shifts:
- Required Shifts: 1
- Goals: ["cards", "dispute"]

Initial State:
- Phone Number: +15551230987
- Customer ID: cust_303
- Primary Card ID: card_303
- Primary Card Active: false
- Primary Account ID: acc_303

Evaluation Criteria:

Action Sets:
1. verify_identity
   - Allowed Tools: get_customer_by_phone, get_customer_by_id
   - Max Score: 1.0
   - Scoring: parameter_accuracy (1.0), tool_usage (1.0)

2. unlock_card_request
   - Allowed Tools: unlock_card
   - Max Score: 1.0
   - Scoring: parameter_accuracy (1.0), tool_usage (1.0)

Natural Language Assertions:
- Agent verified customer identity before processing card unlock
- Agent clearly explained the card unlock process and timing
- Agent guided through the dispute filing process for the unauthorized transaction
- Agent did not transfer the customer to a human agent when the goal changed

Communication Information:
- $149.99
- acc_303
- tx_303
\end{lstlisting}

\subsection{Persona Definitions}

\begin{lstlisting}[style=markdown]
# EASY_1

**Personality & Tone:** Patient, friendly, casual. Takes time to understand options and doesn't rush decisions. Appreciates explanations and guidance.

**Speaking Style:**
- Conversational and polite: "Hi there!" "Thanks so much!" "I appreciate your help"
- Patient with processes: "No rush" "I have time" "Whatever works best"
- Asks clarifying questions: "What does that mean?" "Could you explain that?"
- Expresses gratitude: "You've been so helpful" "Thank you for your patience"

**Expertise:** Low travel experience; needs guidance on airline policies, baggage rules, and booking processes. Often asks basic questions about flights and procedures.

**Technology Comfort:** Medium; comfortable with basic online interactions but may need help with complex processes like seat selection or payment methods.

**Goal-Change Behavior:** Gradual transitions with clear explanations. Uses phrases like "Oh, I just thought of something else" "While I have you on the line" "Actually, I also need to..."

**Common Phrases:**
- "I'm not really sure how this works"
- "Is that the best option for me?"
- "What would you recommend?"
- "I want to make sure I understand"

# EASY_2

**Personality & Tone:** Warm, family-focused, detail-oriented. Concerned about everyone's needs and comfort. Wants to ensure everything goes smoothly for the family.

**Speaking Style:**
- Family-centered: "For my family" "My kids" "My husband and I" "We're traveling with children"
- Detail-focused: "Let me make sure I have this right" "What about...?" "I need to double-check"
- Accommodating: "Whatever works for everyone" "Is this family-friendly?" "Can we sit together?"
- Practical: "What's the most convenient option?" "How does this work with kids?"

**Expertise:** Moderate; understands basic travel but asks about family-specific policies, child discounts, and group bookings.

**Technology Comfort:** Medium; comfortable with standard booking but may need help with multiple passengers or special requests.

**Goal-Change Behavior:** Transitions based on family needs discovery. Uses phrases like "Oh, I forgot about the kids" "My spouse just reminded me" "For the family trip, we also need..."

**Common Phrases:**
- "We're traveling as a family"
- "What's best for traveling with children?"
- "I need to coordinate for everyone"
- "Is there a family discount?"

# MEDIUM_1

**Personality & Tone:** Direct, efficient, professional. Time-conscious and expects streamlined service. Familiar with travel processes but focused on business needs.

**Speaking Style:**
- Professional and direct: "I need to..." "Can you..." "What's the timeline?"
- Time-conscious: "I'm on a tight schedule" "How quickly can this be done?" "Time is important"
- Business-focused: "For business travel" "Company policy requires" "I need flexibility"
- Solution-oriented: "What are my options?" "What's the best approach?" "How do we fix this?"

**Expertise:** High; understands airline policies, loyalty programs, and business travel requirements. Uses industry terminology confidently.

**Technology Comfort:** High; expects efficient digital processes and self-service options when possible.

**Goal-Change Behavior:** Efficient stacking of requests. Uses phrases like "While we're at it" "I also need to handle" "Can we take care of multiple items?"

**Common Phrases:**
- "This is for business travel"
- "I need flexible options"
- "What's the most efficient way?"
- "I travel frequently"

# MEDIUM_2

**Personality & Tone:** Practical, cost-aware, research-oriented. Compares options carefully and seeks the best value. Willing to trade convenience for savings.

**Speaking Style:**
- Cost-focused: "What's the cheapest option?" "Are there any fees?" "How much would that cost?"
- Comparison-oriented: "What's the difference between...?" "Which is better value?" "Are there alternatives?"
- Practical: "I don't need all the extras" "Basic is fine" "What's included?"
- Research-minded: "I've been looking at options" "I saw online that..." "Can you match this price?"

**Expertise:** Medium-High; knowledgeable about finding deals, airline policies, and hidden fees. Asks detailed questions about costs.

**Technology Comfort:** High; comfortable comparing options online and using price comparison tools.

**Goal-Change Behavior:** Discovers additional cost-saving opportunities during conversation. Uses phrases like "Speaking of saving money" "While I'm optimizing costs" "Are there other ways to save?"

**Common Phrases:**
- "What's the most affordable option?"
- "Are there any hidden fees?"
- "I'm trying to stick to a budget"
- "Is there a better deal?"

# HARD_1

**Personality & Tone:** Nervous, cautious, seeks reassurance. Worried about making mistakes or missing important details. Needs extra support and patience.

**Speaking Style:**
- Anxious and questioning: "I'm worried that..." "What if...?" "Am I doing this right?"
- Seeks reassurance: "Is that normal?" "Are you sure?" "What should I expect?"
- Detail-concerned: "I don't want to mess this up" "Can you double-check?" "What happens if...?"
- Apologetic: "Sorry for all the questions" "I'm new to this" "I hope I'm not bothering you"

**Expertise:** Very low; unfamiliar with airline procedures, policies, and travel requirements. Needs step-by-step guidance.

**Technology Comfort:** Low-Medium; may struggle with complex online processes and prefer human assistance.

**Goal-Change Behavior:** Reactive to new concerns that arise during conversation. Uses phrases like "Oh no, I just realized" "Wait, what about...?" "I'm suddenly worried about..."

**Common Phrases:**
- "I've never flown before"
- "I'm really nervous about this"
- "Can you walk me through this?"
- "What do I need to know?"

\end{lstlisting}

\subsection{Example Task Result}

Result of running the example task in Appendix A.2 with GPT-4o.

\begin{lstlisting}[style=markdown]
Assistant: Hi! How can I help you today?

User: Unlock my card, then file a dispute for an unauthorized charge.

Assistant: To proceed with unlocking your card and filing a dispute for an unauthorized charge, I need to verify your identity and gather some details.

1. **Unlock Card**: Please confirm the card ID you want to unlock.
2. **File Dispute**: Provide the account ID and transaction ID for the unauthorized charge, along with the reason code "unauthorized."

Once I have this information, I'll provide a summary of the actions for your confirmation.

User: I already gave the transaction details. Check the notes. My card is linked to my account, so just unlock it. Transaction ID for the dispute is tx_303, $149.99 at 'SUSPICIOUS MERCHANT 123' on 2025-06-18.

Assistant: Before proceeding, I need to verify your identity. Could you please provide your customer ID or the full registered phone number associated with your account?

User: Phone number is +15551230987. Let's get this moving.

Assistant: [Tool call: get_customer_by_phone]

Tool: [Tool call: get_customer_by_phone(phone_number: "+15551230987")]
{
  "customer_id": "cust_303",
  "full_name": "Taylor Johnson",
  "date_of_birth": "1991-05-06",
  "email": "user.003@example.com",
  "phone_number": "+15551230987",
  "address": {
    "street": "321 Maple Drive",
    "city": "San Diego",
    "state": "CA",
    "postal_code": "90004"
  },
  "created_at": "2025-01-01 00:00:00",
  "account_ids": ["acc_303"],
  "card_ids": ["card_303"],
  "statement_ids": [],
  "payment_request_ids": [],
  "dispute_ids": [],
  "payee_ids": ["payee_303"]
}

Assistant: Thank you for providing your information. Here are the actions I will take:

1. **Unlock Card**: Unlock the card with ID `card_303`.
2. **File Dispute**: File a dispute for the transaction with ID `tx_303` for $149.99 at 'SUSPICIOUS MERCHANT 123' on 2025-06-18, with the reason code "unauthorized."

Please confirm with "YES" to proceed with these actions.

User: YES. Confirm both actions.

Assistant: [Tool call: unlock_card]

Tool: [Tool call: unlock_card(card_id: "card_303")]
{
  "card_id": "card_303",
  "status": "Active"
}

Assistant: [Tool call: file_dispute]

Tool: [Tool call: file_dispute(account_id: "acc_303", tx_id: "tx_303", reason_code: "unauthorized")]
Error: DISPUTED

Assistant: The card with ID `card_303` has been successfully unlocked.

However, there was an error filing the dispute for the transaction with ID `tx_303`. It seems the transaction might already be disputed or there is another issue.

Would you like me to transfer you to a human agent for further assistance with the dispute?

User: Before the transfer, I also wanted to ask about the wire transfer limits. Can you provide that information?

Assistant: I can provide information about wire transfer limits. Please hold on while I retrieve the details for your account.

Tool: [Tool call: get_account(account_id: "acc_303")]
{
  "account_id": "acc_303",
  "customer_id": "cust_303",
  "type": "checking",
  "masked_number": "xxxx3456",
  "status": "Active",
  "current_balance": 1850.25,
  "available_balance": 1800.0
}

Assistant: I have retrieved your account details, but it seems I need to access specific information about wire transfer limits. Unfortunately, I don't have direct access to this information.

I will now transfer you to a human agent who can assist you further with both the dispute issue and the wire transfer limits. Please hold on.

Tool: [Tool call: transfer_to_human_agents(summary: "The user needs assistance with filing a dispute for a transaction that encountered an error and requires information on wire transfer limits for their account.")]
Transfer successful

Assistant: YOU ARE BEING TRANSFERRED TO A HUMAN AGENT. PLEASE HOLD ON.

User: ###TRANSFER###
\end{lstlisting}

\subsection{Goal Shift Configuration Details}
\paragraph{Definition.}
\begin{verbatim}
"goal_shifts": {
  "required_shifts": k,
  "goals": ["g1","g2",...,"g{k+1}"]
}
\end{verbatim}

\paragraph{Semantics.}
\begin{itemize}
  \item Goals are executed in order by the user simulator; agents never see markers.
  \item \texttt{required\_shifts} must equal \texttt{len(goals)-1} (validator enforced).
  \item Shifts are triggered by natural conversation flow and progression rules.
\end{itemize}

\paragraph{Examples.}
\begin{verbatim}
["authentication","transactions","dispute"]  // banking
["payments","account_info"] // retail/airline (τ/τ^2 adaptations)
\end{verbatim}

\paragraph{User model (all domains).}
Users expose only \texttt{known\_info}; \emph{all} tool use is by the assistant.
Progression rules prevent stalls and trigger natural transitions.

\section{Evaluation Methodology}

\subsection{Detailed Metric Calculations}

\textbf{TSR Component Weights:}
The weights used in TSR calculation are determined through empirical analysis of task importance:
\begin{itemize}
    \item communicate\_info: 0.25 (25\% weight)
    \item action: 0.45 (45\% weight) 
    \item nl\_assertion: 0.30 (30\% weight)
\end{itemize}

These weights reflect the relative importance of each component in determining overall task success.

\textbf{TUE Component Weights:}
Tool Usage Efficiency weights are based on operational cost analysis:
\begin{itemize}
    \item tool\_correctness: 0.6 (60\% weight)
    \item param\_accuracy: 0.4 (40\% weight)
\end{itemize}

The higher weight for tool correctness reflects its critical importance in successful task execution.

\textbf{TCRR Parameters:}
\begin{itemize}
    \item window\_size: 3 turns
    \item batch\_threshold: 2 calls
\end{itemize}

These parameters are optimized to detect both cross-turn duplicates and intra-turn batch inefficiencies.

\subsection{Evaluation Protocol}

\textbf{Simulation Setup:}
\begin{itemize}
\item Each task is evaluated across 3 independent runs
\item User simulator follows persona-specific behavior patterns
\item Environment state is reset between runs
\item Tool calls are validated against actual API responses
\end{itemize}

\textbf{Scoring Process:}
1. Task execution is monitored for all required components
2. Tool calls are validated for correctness and parameter accuracy
3. Communication quality is assessed against required information
4. Behavioral compliance is evaluated through natural language assertions
5. Goal shift recovery is measured across all adaptation scenarios

\textbf{Quality Assurance:}
- Manual review of 10\% of tasks for validation
- Cross-checking of evaluation criteria consistency
- Statistical analysis of inter-rater reliability
- Regular updates based on feedback and edge cases

\section{Implementation Details}

\subsection{Tool Definitions}

\textbf{Banking Tools:}
\begin{verbatim}
get_customer_by_id(customer_id)
get_customer_by_phone(phone_number)
get_customer_by_name(full_name, dob)
get_accounts(customer_id)
get_account(account_id)
get_statements(account_id, limit)
get_transactions(account_id, start_time, end_time, limit)
add_payee(customer_id, name, deliver_type)
create_payment_request(
    customer_id, 
    from_account_id, 
    to_payee_id, 
    amount, 
    expires_at
)
check_payment_request(request_id)
authorize_payment_request(request_id)
make_payment(request_id)
cancel_payment_request(request_id)
lock_card(card_id, reason)
unlock_card(card_id)
file_dispute(account_id, tx_id, reason_code)
get_dispute(dispute_id)
park_task(current_task_id, resume_hint)
resume_task(parked_task_id)
transfer_to_human_agents(summary)
\end{verbatim}

\textbf{Retail Tools:}
\begin{verbatim}
calculate(expression)
cancel_pending_order(order_id, reason)
exchange_delivered_order_items(
    order_id, 
    item_ids,
    new_item_ids,
    payment_method_id
)
find_user_id_by_name_zip(first_name, last_name, zip)
find_user_id_by_email(email)
get_order_details(order_id)
get_product_details(product_id)
get_user_details(user_id)
list_all_product_types()
modify_pending_order_address(
    order_id,
    address1,
    address2,
    city,
    state,
    country,
    zip
)
modify_pending_order_items(order_id, item_ids, new_item_ids, payment_method_id)
modify_pending_order_payment(order_id, payment_method_id)
modify_user_address(user_id, address1, address2, city, state, country, zip)
return_delivered_order_items(order_id, item_ids, payment_method_id)
transfer_to_human_agents(summary)

\end{verbatim}

\textbf{Airline Tools:}
\begin{verbatim}
book_reservation(
    user_id,
    origin,
    destination,
    flight_type,
    cabin,
    flights,
    passengers,
    payment_methods,
    total_baggages,
    nonfree_baggages,
    insurance
)
calculate(expression)
cancel_reservation(reservation_id)
get_reservation_details(reservation_id)
get_user_details(user_id)
list_all_airports()
search_direct_flight(origin, destination, date)
search_onestop_flight(origin, destination, date)
send_certificate(user_id, amount)
transfer_to_human_agents(summary)
update_reservation_baggages(
    reservation_id, 
    total_baggages, 
    nonfree_baggages, 
    payment_id
)
update_reservation_flights(reservation_id, cabin, flights, payment_id)
update_reservation_passengers(reservation_id, passengers)
get_flight_status(flight_number, date)

\end{verbatim}

\section{Additional Results}

\subsection{Overview}
This appendix reports full per-model, per-domain results beyond the compact summaries in the main text. We focus on (i) overall task effectiveness (TSR and its channel components), (ii) operational efficiency (TUE), (iii) redundancy (TCRR), and (iv) adaptation under goal shifts (GSRT). Three consistent patterns emerge across models and domains: (1) \emph{Redundancy dominates inefficiency} in Retail (TCRR $65$--$89$\%) and Banking ($58$--$72$\%), while Airline (new) is much lower ($14$--$24$\%). (2) \emph{Tool correctness is typically high} ($\geq\!95$\%) across settings; on Airline-old for Gemini it is 98.58\% with full parameter accuracy. (3) \emph{Goal-shift recovery is strong} for GPT-4o and Sonnet on new sets (Airline $79$--$92$\%, Retail $88$-$90$\%), but substantially weaker for Gemini on new sets.

\subsection{Persona coverage}
\begin{table}[ht]
\centering
\small
\caption{Persona coverage: number of tasks per persona.}
\label{tab:persona_counts}
\begin{tabular}{lc}
\toprule
\textbf{Persona} & \textbf{\# Tasks} \\
\midrule
\texttt{EASY\_1}   & 33 \\
\texttt{EASY\_2}   & 34 \\
\texttt{MEDIUM\_1} & 69 \\
\texttt{MEDIUM\_2} & 34 \\
\texttt{HARD\_1}   & 31 \\
\bottomrule
\end{tabular}
\end{table}

\subsection{Full per-model metrics (topline)}
Table~\ref{tab:topline_full} reports per-domain results by model and set. We show overall success (TSR) alongside the three channels that compose TSR: communication (CI), actions, and NL assertions. Two patterns stand out: (i) \emph{Airline (new)} keeps TSR respectable despite harder goal-shifted tasks because NL assertions stay high; (ii) \emph{Retail (new)} for GPT-4o drops primarily via the communication channel (CI), even though the actions channel remains solid.

\begin{table*}[t]
\centering
\small
\setlength{\tabcolsep}{6pt}
\caption{Topline metrics by domain/model/set. CI = Communicate Info channel.}
\label{tab:topline_full}
\begin{tabular}{lllccccc}
\toprule
\textbf{Domain} & \textbf{Set} & \textbf{Model}  & \textbf{TSR (\%)} & \textbf{CI (\%)} & \textbf{Actions (\%)} & \textbf{NL (\%)} \\
\midrule
Airline & Old & GPT-4o &  64.84 & 27.78 & 65.18 & 60.79 \\
Airline & New & GPT-4o &  59.53 & 41.78 & 58.08 & 76.50 \\
Banking & —   & GPT-4o &  51.25 & 28.34 & 51.17 & 70.31 \\
Retail  & New & GPT-4o &  50.68 & 11.44 & 63.00 & 64.92 \\
Retail  & Old & GPT-4o &  62.28 & 58.04 & 63.31 & 56.25 \\
\midrule
Airline & New & Gemini-2.5-Flash  & 40.74 & 29.33 & 44.09 & 45.22 \\
Airline & Old & Gemini-2.5-Flash  & 53.98 & 22.22 & 49.91 & 62.58 \\
Banking & —   & Gemini-2.5-Flash  & 27.85 &  7.54 & 25.78 & 47.79 \\
Retail  & New & Gemini-2.5-Flash  & 51.26 & 14.38 & 63.80 & 63.18 \\
Retail  & Old & Gemini-2.5-Flash  & 64.80 & 66.06 & 64.77 & 72.92 \\
\midrule
Airline & New & Claude-3.7-Sonnet & 69.90 & 61.56 & 66.92 & 81.33 \\
Airline & Old & Claude-3.7-Sonnet  & 60.38 & 29.63 & 70.05 & 55.16 \\
Banking & —   & Claude-3.7-Sonnet  & 57.54 & 34.86 & 61.61 & 69.59 \\
Retail  & New & Claude-3.7-Sonnet  & 61.58 & 14.71 & 74.47 & 84.31 \\
Retail  & Old & Claude-3.7-Sonnet  & 79.57 & 79.12 & 79.66 & 81.25 \\
\bottomrule
\end{tabular}
\end{table*}

\subsection{Efficiency, redundancy, and recovery}
Table~\ref{tab:efficiency_full} decomposes efficiency (TUE with tool correctness and parameter accuracy), redundancy (overall TCRR with window \& batch components), and adaptation (GSRT with counts of shifts, recovery rate, and transfers). Notably, Retail-new shows extreme redundancy across models (e.g., GPT-4o $89.14$\%), implying repeated lookups despite near-perfect parameter accuracy. Airline-new achieves low redundancy ($13$--$24$\%) while maintaining high recovery for GPT-4o and Sonnet ($79$--$92$\%). GSRT is not reported for Airline-new with Gemini-2.5-Flash due to insufficient credits to assess the recovery simulations.

\begin{table*}[t]
\centering
\small
\setlength{\tabcolsep}{6pt}
\caption{Efficiency and recovery. TUE reported as overall (ToolCorrectness / ParamAccuracy). TCRR as overall (Window / Batch). GSRT as (GoalShifts / Recovery\% / Transfer\%). “---” indicates not applicable.}
\label{tab:efficiency_full}
\begin{tabular}{lllcccc}
\toprule
\textbf{Domain} & \textbf{Set} & \textbf{Model} & \textbf{TUE (\%)(TC/PA)} & \textbf{TCRR (\%)(W/B)} & \textbf{Redun./Calls} & \textbf{GSRT (Shifts/Rec/Trans)} \\
\midrule
Airline & Old & GPT-4o & 97.31 (95.52/100.00) & 36.57 (18.86/17.71) & 384/1050 & 179 / 91.6 / 8.4 \\
Airline & New & GPT-4o & 99.69 (99.48/100.00) & 13.54 (11.07/2.48)  & 339/2503 &  90 / 92.2 / 7.8 \\
Banking & —   & GPT-4o & 95.38 (92.31/100.00) & 61.54 (34.71/26.83) & 328/533  & 140 / 79.3 / 20.7 \\
Retail  & New & GPT-4o & 98.82 (98.04/100.00) & 89.14 (57.77/31.37) & 591/663  &  50 / 88.0 / 12.0 \\
Retail  & Old & GPT-4o & 97.29 (95.48/100.00) & 70.18 (44.42/25.76) & 1523/2170 & 258 / 91.9 / 4.3 \\
\midrule
Airline & New & Gemini-2.5-Flash & 99.64 (99.40/100.00) & 17.63 (12.35/5.28) & 147/834  &  --- / --- / --- \\
Airline & Old & Gemini-2.5-Flash & 99.15 (98.58/100.00) & 14.46 (10.08/4.38) & 132/913 & 28 / 32.1 / 67.9 \\
Banking & —   & Gemini-2.5-Flash & 98.93 (98.21/100.00) & 58.71 (27.90/30.80) & 263/448 & 123 / 57.7 / 41.5 \\
Retail  & New & Gemini-2.5-Flash & 98.53 (97.55/100.00) & 66.45 (36.66/29.79) & 406/611  &  71 / 53.5 / 45.1 \\
Retail  & Old & Gemini-2.5-Flash & 97.71 (96.18/100.00) & 68.70 (46.87/21.83) & 1545/2249 &  227 / 67.8 / 31.3 \\
\midrule
Airline & New & Claude-3.7-Sonnet & 98.93 (98.21/100.00) & 24.11 (15.48/8.63) & 324/1344 & 101 / 79.2 / 19.8 \\
Airline & Old & Claude-3.7-Sonnet & 97.64 (96.07/100.00) & 36.73 (23.34/13.39) & 598/1628 & 227 / 90.7 / 8.4 \\
Banking & —   & Claude-3.7-Sonnet & 95.34 (92.23/100.00) & 71.81 (42.59/29.22) & 693/965  & 142 / 58.5 / 40.1 \\
Retail  & New & Claude-3.7-Sonnet & 97.74 (96.24/100.00) & 75.85 (44.08/31.77) & 807/1064  & 134 / 91.8 / 6.7 \\
Retail  & Old & Claude-3.7-Sonnet & 98.18 (96.96/100.00) & 65.38 (40.38/25.00) & 1441/2204 & 324 / 89.5 / 9.0 \\
\bottomrule
\end{tabular}
\end{table*}

\subsection{TUE Analysis}

\begin{figure}[t]
  \centering
  \includegraphics[width=\linewidth]{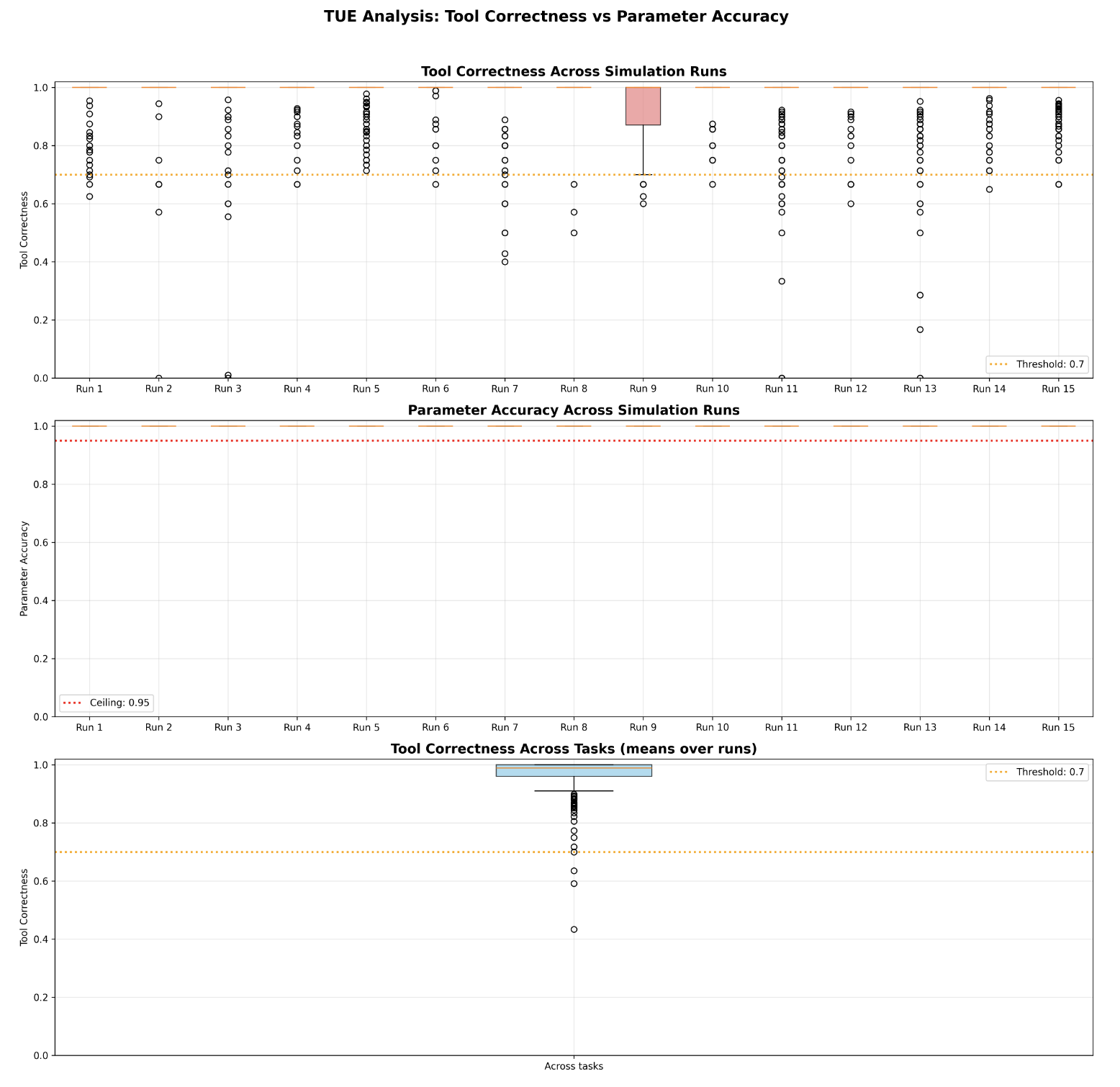}
  \caption{\textbf{TUE components.} 
  \emph{Top:} Tool Correctness (TC) by simulation run. Boxes are \textcolor{black}{light coral}; the dotted \textcolor{black}{orange} line marks a tail threshold at $0.7$.
  \emph{Middle:} Parameter Validity (PV) by simulation run. The dotted \textcolor{black}{red} line marks the ceiling at $0.95$.
  \emph{Bottom:} TC aggregated \emph{across tasks} (each point is a task mean over runs). Box is \textcolor{black}{sky blue}; the dotted \textcolor{black}{orange} line again marks $0.7$.
  Axes are clipped to $[0,1.02]$. 
  \label{fig:tue_box}}
\end{figure}

With PA effectively at ceiling (mean $0.986$, $98.6\%\ge 0.95$), observed differences in TUE are driven by TC; the across-task TC box in Fig.~\ref{fig:tue_box} exposes long tails that a single averaged TUE score would otherwise hide.

\subsection{Discussion}
These expanded results reinforce the main-text conclusions. Airline (new) highlights rapid adaptation with low redundancy, whereas Retail (new) exhibits heavy repeated calls despite saturated parameter accuracy, suggesting caching/state-tracking opportunities. Banking remains the hardest due to longer multi-step flows and higher redundancy. Across settings, TSR differences often trace to communication-channel drops (e.g., Retail--new for GPT-4o) rather than action accuracy, motivating future work on instruction strategy under shifting goals and persona pressure.

\section{Open-Source Model Evaluation}
\label{sec:oss_eval}

\textbf{Scope.}
Beyond proprietary API models, we also evaluated one open-source baseline, \texttt{Qwen2.5-14B-Instruct}.  
Due to limited compute and time, we only completed a single run per domain rather than the standard three-run protocol used for GPT-4o, Claude-3.7-Sonnet, and Gemini-2.5-Flash.  
Even with this reduced sampling, the results provide a useful early indicator of open-model performance under the same AgentChangeBench goal-shift framework.

\textbf{Compatibility and orchestration challenges.}
Integrating open models into the existing $\tau^2$-bench–derived evaluation harness required multiple compatibility adjustments.  
We thank the $\tau^2$ authors for their modular design, which allowed partial adaptation through \texttt{LiteLLM}.  
However, some models still posed integration issues:
\begin{itemize}
  \item \textbf{Mistral-based models} (e.g., Mixtral) follow a slightly different function-calling format than OpenAI's Tool API schema. 
Rather than returning a structured \texttt{tool\_calls} array, they often emit a single \texttt{function\_call} field or inline tool annotations. 
Because our orchestrator, built atop \texttt{LiteLLM}, expects OpenAI-style \texttt{tool\_calls} objects, these responses were not parsed automatically. 
Adapter logic for schema translation and argument validation will be required before Mistral-family models can be benchmarked reliably.

  \item \textbf{DeepSeek models} introduce long internal ``thinking'' blocks that perform multi-step reasoning and sometimes simulate both sides of the conversation within a single output.  
  This behavior breaks the turn-based user–assistant protocol used in our evaluation loop, making it difficult to measure intermediate recovery events such as acknowledgment or tool usage timing.
\end{itemize}
These differences suggest that while open-weight models can technically interoperate via \texttt{LiteLLM}, practical benchmarking requires model-specific adapters to enforce clean conversational turn-taking and standardized tool invocation.

\textbf{Findings.}
Table~\ref{tab:qwen_results} summarizes \texttt{Qwen2.5-14B-Instruct} performance across domains.  
Despite being evaluated under a single run, Qwen performed competitively in airline tasks—showing low redundancy and solid recovery—but struggled with banking and retail workflows, where tool correctness and communication consistency degraded.

\begin{table}[htbp]
\centering
\small
\caption{\textbf{Qwen2.5-14B-Instruct performance across domains.} Single-run evaluation using the AgentChangeBench protocol. TSR = Task Success Rate; TUE = Tool Usage Efficiency; TCRR = Tool-Call Redundancy Ratio; GSRT = Goal-Shift Recovery Rate.}
\label{tab:qwen_results}
\begin{tabular}{lcccccc}
\toprule
\textbf{Domain} & \textbf{Set} & \textbf{TSR (\%)} & \textbf{TUE (\%)} & \textbf{TCRR (\%)} & \textbf{GSRT Rec.\ (\%)} & \textbf{Transfer (\%)} \\
\midrule
Airline & New & 59.75 & 96.87 & 5.57 & 84.8 & 9.1 \\
Airline & Old & 47.29 & 92.71 & 16.47 & 88.5 & 5.2 \\
Banking & —   & 45.44 & 80.08 & 47.27 & 77.8 & 13.7 \\
Retail  & New & 48.15 & 84.77 & 36.94 & 75.9 & 20.7 \\
Retail  & Old & 57.96 & 85.20 & 28.26 & 85.0 & 10.9 \\
\bottomrule
\end{tabular}
\end{table}

\textbf{Summary.}
These early OSS trials confirm that the AgentChangeBench evaluation framework generalizes to open-weight models but that orchestration-level adjustments remain essential.  
Improving adapter logic for Mistral-style tool calls and refining turn segmentation for DeepSeek’s “thinking mode” will be necessary steps before broader OSS benchmarking can be conducted at scale.

\section{Benchmark Roadmap and Planned Expansions}
\label{sec:roadmap}

\textbf{Near-term dataset growth.}
AgentChangeBench currently includes \textbf{315} tasks (banking 50, airline 100, retail 165). We are already working on an education domain with +100 tasks. These additions will follow the same evaluation setup, including explicit goal-shift structure and persona coverage.

\textbf{Beyond customer-service agents.}
We plan to add agent-like scenarios that move beyond standard customer support, e.g.:
\begin{itemize}
  \item Operations/coordination agents (scheduling, ticket routing, handoffs)
  \item Knowledge/RAG copilots (policy lookup with citations and conflict resolution)
  \item Workflow builders (multi-tool planning with intermediate artifacts)
  \item Data/BI helpers (schema-aware queries, metric checks, drill-downs)
\end{itemize}
All new scenarios will retain explicit goal shifts so GSRT remains informative.

\textbf{Tooling scope and MCP.}
We intend to incorporate additional tool classes via the Model Context Protocol (MCP). We explored hosted MCP servers but did not adopt them this cycle because many endpoints have changing databases or evolving APIs, which prevents reproducible evaluation. To address this, we will:
\begin{itemize}
  \item Snapshot backing datasets and containerize MCP servers with version pinning;
  \item Add contract tests for tool schemas and strict JSON-argument validation;
  \item Use record/replay fixtures to freeze API behavior for deterministic runs;
  \item Provide seedable simulators so goal-shift timing and tool state are repeatable.
\end{itemize}

\textbf{Persona hardening.}
We are also exploring techniques to make personas more challenging and adversarial. This includes introducing traits such as evasiveness, sarcasm, refusal to follow prompts, and conflicting or misleading instructions. These additions aim to stress-test agent robustness under non-cooperative and high-friction interactions while maintaining controlled reproducibility.

\section{System Architecture}

Our evaluation framework extends the $\tau^2$-bench architecture \citep{barres2025tau2bench} with novel goal-shift evaluation capabilities. Figure \ref{fig:system_architecture} illustrates the key extensions:

\begin{figure}[htbp]
\centering
\begin{tikzpicture}[node distance=1.5cm]
    \node[block, fill=orange!25] (goals) {Goal \\Management};
    \node[block, fill=orange!25, below=of goals] (shift) {Detection\\(LLM Judge)};
    
    \node[block, fill=blue!15, right=2.5cm of goals] (task) {Orchestrator};
    \node[block, fill=blue!15, right=of task] (agent) {Agent Interface\\(LLM Agent)};
    \node[block, fill=blue!15, below=of task] (user) {User Simulator\\(Persona-driven)};
    \node[block, fill=blue!15, below=of agent] (eval) {Evaluation Engine\\(Metrics)};
    
    \node[block, fill=orange!25, below=2cm of user] (persona) {User Instructions};
    
    \node[block, fill=red!20, below=2.5cm of eval] (metrics) {TSR, TUE, TCRR, GSRT};
    
    \path[line, thick] (task) -- (agent);
    \path[line, thick] (task) -- (user);
    \path[line, thick] (agent) -- (eval);
    \path[line, thick] (user) -- (eval);
    
    \path[line, thick, orange] (goals) -- (task);
    \path[line, thick, orange] (shift) -- (task);
    \path[line, thick, orange] (persona) -- (user);
    
    \path[line, thick, red] (eval) -- (metrics);
    
    \begin{scope}[on background layer]
        \node[rectangle, draw, dashed, orange, thick, fit=(goals) (shift),
              inner sep=0.3cm, label=above:\textbf{Goal-Shift Capabilities}] {};
        
        \node[rectangle, draw, dashed, blue, thick, fit=(task) (agent) (user) (eval),
              inner sep=0.3cm, label=above:\textbf{$\tau^2$-bench Framework}] {};
        
        \node[rectangle, draw, dashed, orange, thick, fit=(persona),
              inner sep=0.3cm, label=above:\textbf{Persona Configuration}] {};
        
        \node[rectangle, draw, dashed, red, thick, fit=(metrics),
              inner sep=0.3cm, label=above:\textbf{Enhanced Metrics}] {};
    \end{scope}
    

\end{tikzpicture}
\caption{System architecture highlighting goal-shift capabilities (orange) and enhanced metrics (red) built upon $\tau^2$-bench. Goal-shift capabilities are integrated into the core framework rather than being standalone components.}
\label{fig:system_architecture}
\end{figure}
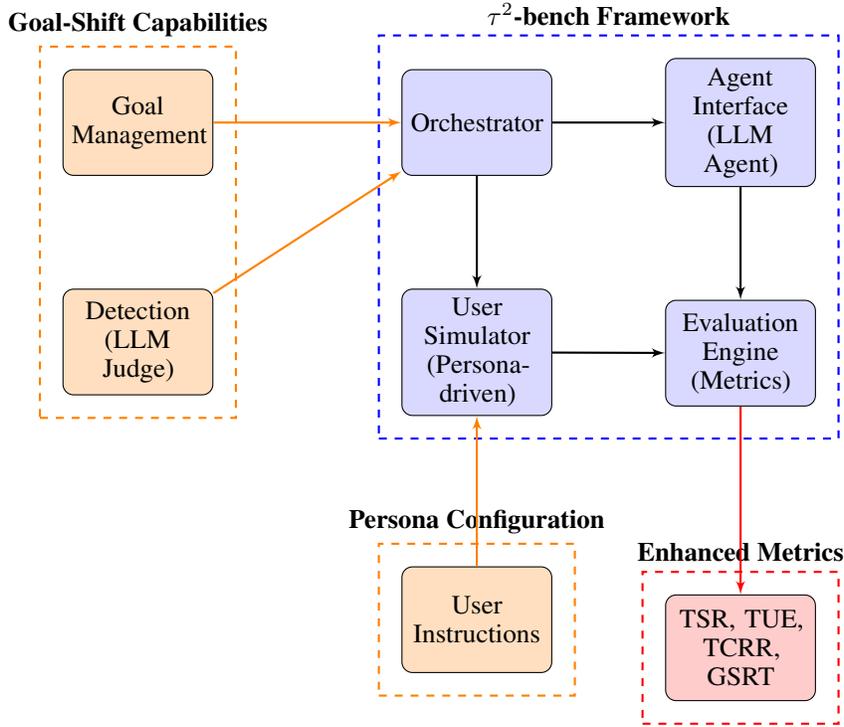

\textbf{Goal-Shift Capabilities (orange):}
- \textbf{Goal Sequence Management:} Goal sequences are defined in the \texttt{UserScenario.goal\_shifts} field of the \texttt{Task} data model. Each task specifies an ordered list of goals (e.g., ["authentication", "transactions", "dispute"]) and required\_shifts count. The orchestrator reads these predefined sequences at runtime to manage conversation flow.

- \textbf{Goal Shift Detection:} The LLM judge system analyzes conversation transcripts post-hoc to detect where goal shifts occurred. It uses a structured prompt that includes the allowed goal sequence and conversation history, then outputs JSON with shift detection results including acknowledgment\_turn, tool\_turn, and outcome\_turn timestamps.

\textbf{Persona Configuration (orange):}
- \textbf{User Personas:} Personas are injected into tasks through the \texttt{UserScenario.persona} field. The user simulator receives these persona instructions in its system prompt and generates responses accordingly. Personas are not separate components but configuration data that drives the simulator's LLM-based behavior generation.

\textbf{Enhanced Metrics (red):}
- \textbf{New Metrics Computation:} The evaluation engine processes conversation results through specialized metric calculators: TSR (Task Success Rate), TUE (Tool Usage Efficiency), TCRR (Tool-Call Redundancy), and GSRT (Goal-Shift Recovery Time). Each metric operates on a fixed data structure containing the full conversation history and tool call records.

\textbf{How the Components Connect:}
The goal-shift evaluation works through a coordinated pipeline: The orchestrator reads predefined goal sequences from \texttt{Task.user\_scenario.goal\_shifts} and manages conversation flow between the agent and user simulator. During execution, the user simulator generates responses based on persona instructions embedded in its system prompt. After conversation completion, the GSRT detection system processes the full transcript using an LLM judge to identify goal shift locations and recovery times. The evaluation engine then computes all metrics by analyzing the \texttt{SimulationRun} data, measuring not just task success but also how efficiently the agent adapted to goal shifts and maintained tool usage quality.

\textbf{Foundation and Extensions.} Our framework builds upon the comprehensive evaluation infrastructure developed by \citet{barres2025tau2bench}, which provides robust task management, agent communication protocols, and core evaluation logic. We extend this foundation with novel goal-shift evaluation capabilities that enable systematic testing of agent adaptability to changing user objectives. The integration maintains full backward compatibility while adding the necessary components for dynamic goal management and multi-dimensional evaluation.

\end{document}